\documentclass[conference]{IEEEtran}
\IEEEoverridecommandlockouts
\usepackage{cite}
\usepackage[numbers]{natbib}
\usepackage{multicol}
\usepackage[bookmarks=true]{hyperref}
\usepackage{graphics} 
\usepackage{epsfig} 
\usepackage{mathptmx} 
\usepackage{times} 
\usepackage{amsmath} 
\usepackage{amssymb}  
\graphicspath{ {./images/} }
\usepackage{balance}
\usepackage{float}
\usepackage{textcomp}
\usepackage{algorithm}
\usepackage{algorithmic}
\usepackage{graphicx}
\usepackage{caption}
\usepackage{subcaption}
\usepackage{color,soul}
\usepackage{bm}
\usepackage[utf8]{inputenc}

\usepackage{hyperref}

\usepackage{booktabs}
\usepackage{array}
\usepackage{graphics} 
\usepackage{epsfig} 
\usepackage{amsmath} 
\usepackage[section]{placeins}
\usepackage{multirow}
\usepackage{xcolor}
\usepackage{subcaption}
\usepackage[utf8]{inputenc}
\usepackage{amsmath}

\def\BibTeX{{\rm B\kern-.05em{\sc i\kern-.025em b}\kern-.08em
    T\kern-.1667em\lower.7ex\hbox{E}\kern-.125emX}}
\ifCLASSINFOpdf
\else
\fi
\begin{document}
%
\title{RoboMal: Malware Detection for Robot Network Systems}



%
\author{\IEEEauthorblockN{Upinder Kaur\IEEEauthorrefmark{1},
Haozhe Zhou\IEEEauthorrefmark{2},
Xiaxin Shen\IEEEauthorrefmark{3}, 
Byung-Cheol Min\IEEEauthorrefmark{3} and
Richard M. Voyles\IEEEauthorrefmark{1}}
\IEEEauthorblockA{\IEEEauthorrefmark{1}School of Engineering Technology,\\}
\IEEEauthorblockA{\IEEEauthorrefmark{2}Department of Computer Science,\\}
\IEEEauthorblockA{\IEEEauthorrefmark{3}Department of Computer and Information Technology,\\ Purdue University,\\
West Lafayette, IN 47907, USA\\ Email: kauru, zhou929, shen452, minb, rvoyles@purdue.edu}
}

\IEEEspecialpapernotice{Accepted to IRC 2021}

\maketitle

\begin{abstract}
Robot systems are increasingly integrating into numerous avenues of modern life. From cleaning houses to providing guidance and emotional support, robots now work directly with humans. Due to their far-reaching applications and progressively complex architecture, they are being targeted by adversarial attacks such as sensor-actuator attacks, data spoofing, malware, and network intrusion. Therefore, security for robotic systems has become crucial. In this paper, we address the under-served area of malware detection in robotic software. Since robots work in close proximity to humans, often with direct interactions, malware could have life-threatening impacts. Hence, we propose the RoboMal framework of static malware detection on binary executables to detect malware before it gets a chance to execute. Additionally, we address the great paucity of data in this space by providing the RoboMal dataset\footnote{The RoboMal dataset is available at \url{https://doi.org/10.4231/YN7G-H807}.} comprising controller executables of a small-scale autonomous car. The performance of the framework is compared against widely used supervised learning models: GRU, CNN, and ANN. Notably, the LSTM-based RoboMal model outperforms the other models with an accuracy of 85\% and precision of 87\% in 10-fold cross-validation, hence proving the effectiveness of the proposed framework.
\end{abstract}



%
\IEEEpeerreviewmaketitle

\section{Introduction}\label{sec:intro}
Robotic systems, much like other cyber physical systems (CPS), are becoming increasingly integrated into modern life. Robots are no longer limited to strict research or industrial setting but are finding space in offices, homes, and schools. Robots with enhanced sensing such as computer vision, audio recognition, and autonomous decision-making capabilities are making everyday life easier and safer for humans. From autonomous cars to home cleaning, robots are shifting the paradigm of modern living. 

Due to their increasing adoption, robotic systems are becoming an attractive target of malicious attacks \cite{morante2015cryptobotics}. From interruption of services, to falsification and theft of data, increasingly sophisticated methods are being deployed to attack various components of a robotic system \cite{rescuerobots,gomez2016hardware, maliciousroboticsystems}. Hence, security for robotic systems has become an ever-evolving challenge encompassing physical as well as communication and software layers of the architecture. 

Attacks on robotic systems target its core properties: confidentiality, integrity, and availability \cite{geris2019feature}. Confidentiality attacks include spoofing, unauthorized access, and password pilfering; whereas tampering, malware, message relay, data injection, and data modification attack the integrity of the system \cite{rouzbahani2020anomaly}. Further, a critical aspect of attacks on robotic systems could be the intention to cause direct physical harm to humans. Robots today work in close proximity with humans with direct interactions. An attacker that can control such a robot can potentially have life-threatening impacts \cite{chung2019smart, dash2019out}.  

Malware attacks include injection of software which forces the system to misbehave \cite{sharmeen2019identifying}. They can be injected by infiltrating the network or attacking nodes such as edge devices or fog nodes. The key aspect of malware is its close resemblance to actual good software that makes detection hard \cite{brand2010malware}. A good example of this is Stuxnet \cite{marr2019cyberwarfare}, a malware that went undetected for years and threatened the nuclear capability of a country. This piece of code was not detected by anti-virus software due to its similarity with usual good windows software. This is even more critical for low-level embedded devices such as those used in robotic systems. For example, by just adjusting the gains, a Proportional–Integral–Derivative (PID) controller can be made to misbehave entirely, thereby threatening the integrity of the system. Since these changes are so minute, they are challenging to detect by traditional approaches.

Malware attacks not only threaten information security, but since CPS involves physical actuation, they can cause real-world life-threatening damage. For example, misbehaving surgical robots \cite{bonaci2015make}, bricking autonomous cars \cite{dash2019out}, causing disruption on a manufacturing floor with robots \cite{maggi2017rogue}, are few such attack vectors that have been studied. Hence, detection and subsequent isolation of malware are critical for robust and safe robotic systems. Moreover, due to the evolving and mimicking nature of malware, traditional rule-based approaches prove to be unsuitable \cite{anderson2018ember}. Further, traditional seed-key based security is often too expensive and impractical to implement in CPS devices due to their constrained resources. Therefore, learning-based approaches can be a potential solution that can be independently deployed on each device. Also, new innovations in lighter, more optimized learning models for embedded systems (e.g., TinyML, Tensorflow Lite \cite{tinyml}) further strengthens the case for using machine learning. 

In this paper, we address this need for malware detection in robotic systems by presenting a learning-enabled framework. The aim of this work is to detect malware before it even has a chance to install or execute, thereby greatly limiting the impact that malware can have. While such static malware detection has been studied for personal computers and android-based devices, the area is under-served for low-level embedded devices. Yet, as more robots are now connected to networks, the need is for fast and optimized malware detection attuned for the limited resources of embedded devices.


The contributions of this paper are:
\begin{itemize}
    \item The RoboMal dataset which comprises binary executables of controllers of a small-scale multi-sensor autonomous car. This dataset is made available to the public for fostering further research in the area.
    \item The RoboMal framework which processes sequential information at the byte level to detect malware in executables of the robot software. 
    \item A detailed comparison and analysis of the performance of the proposed framework along with other state-of-the-art models on the RoboMal dataset. 
\end{itemize}

The organization of this paper is as follows: Section \ref{sec:mcps} describes the previous works in malware detection in robot systems. The RoboMal dataset is described in Section \ref{sec:dataset}. The proposed RoboMal framework for malware detection is presented in Section \ref{sec:model}. The experiments and results are presented in Section \ref{sec:results}. Section \ref{sec:con} presents the conclusion of this work along with its future scope. 
\section{Malware in Robot Networks}\label{sec:mcps}
Robot systems can be attacked through multiple modalities. While numerous works have been published for sensor-actuator attacks \cite{gao2020state, sa2}, data spoofing \cite{cardenas2009challenges}, and fraud and intrusion detection \cite{10.1145/2542049,intrusiondetection} which manipulate data being exchanged, few have addressed malware detection in robotics. Malicious software, or malware, is an ever-evolving threat faced by most computing systems today \cite{8405951}. Anti-virus companies detect malware by matching unseen code against massive libraries of identified malware. While this approach works for most malware, it often proves to be an expensive endeavor due to the high cost of updating and maintaining such libraries. Further, as seen in recent times, malware attacks have become increasingly sophisticated, often outmatching traditional malware detection systems \cite{10.1145/3230833.3233280}. Robot systems are particularly vulnerable to malware not only due to the paucity of available protections, but also due to the fact that even a change of a single byte can completely change the behavior of a robot \cite{chung2019smart}. As more and more robots come online, connect to large networks, there is an imminent need for malware detection and isolation in robotic systems.

\begin{figure}
  \centering
  \includegraphics[width=0.9\linewidth]{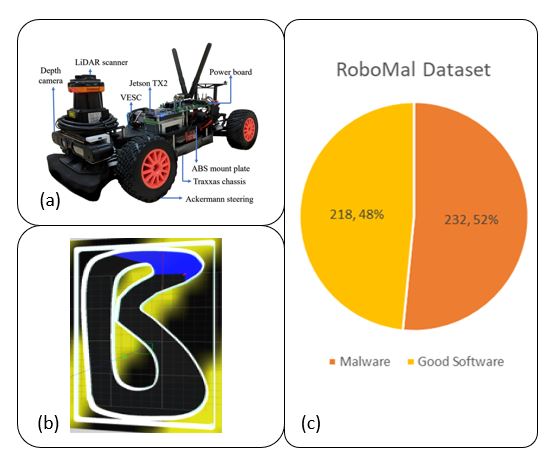}
  \caption{The RoboMal Dataset; (a) The F1/10 autonomous car with multiple sensors \cite{o2019f110}, (b) The wall follower track in Gazebo simulator, and (c) The distribution of malware and good software in RoboMal dataset.}
  \label{fig:robomal_car_datset}
\end{figure}

In this work, we consider the threat model of malicious code executables being communicated to an autonomous robot. Such code is often sent for remote updates and due to limited protection, it can be spoofed to inject malware. Autonomous robots often do not have the resources to afford traditional encryption-based security. Hence, this makes them vulnerable to malware. Malware detection in computer systems has seen numerous works using learning-based approaches to identify maliciousness in code as such models can evolve with time. Learning-based malware detection systems are of two main types - dynamic and static. In dynamic analytical systems, runtime behaviors such as API calls, execution behavior, instructions, etc. are analyzed to classify programs as malware \cite{journals/jcs/RieckTWH11,stokes}. This approach captures behavioral patterns and while it may seem more intuitive to develop, it demands an isolated environment for the identification of behavior. Customized virtual machine testbeds needed for such analysis are not only expensive but accrue significant computational costs. Also, intelligent malware might have the ability to recognize such testbeds and could potentially avoid discovery \cite{raffetseder2007detecting}.  

Static analysis based on supervised approaches using neural networks on portable executables (PE) has shown better performance \cite{saxe2015,10.1145/raff2017,raff2017malware,rvinay}. Deep learning on metadata features, contextual byte features, and 2D histograms from the PE files resulted in a detection rate of 95\% \cite{saxe2015}. Long Short-term Memory (LSTM) models have also been used to learn from domain knowledge extracted using n-gram methods \cite{raff2017malware}. This work was further extended using convolutional neural networks (CNN) with temporal max-pooling in the MalConv model \cite{10.1145/raff2017} which resulted in state-of-the-art performance. Moreover, visualization techniques that convert binary executable into images for identifying malware have also been attempted  \cite{10.1145/malwareimage1,venkatraman2018use,venkatraman2019hybrid}. Such systems employed CNN and deep neural networks to identify textural patterns in malware images. Although these methods achieved high detection rates, they still rely on large labeled datasets. 

In embedded systems, malware detection has been studied for Android systems \cite{6298824,8629067} wherein system calls and two-dimensional features were used, respectively. While these approaches have shown good precision, they fail to work at the granularity needed for robotic code. The features used in these methods result in too much information loss for small embedded code files written for robotic systems. Further, they rely on massive datasets (for example, the Embers dataset has 200,000 samples) that do not yet exist for robotic systems. Moreover, they are constrained to the environment such as windows or android and do not take into account the generalizability needed for robotic software. Hence, for robot systems, the need is to examine the code with greater granularity to capture the finest of details. Further, the approach needs to be language agnostic as robotic code can be developed in various environments and systems using different tools and languages.

\section{The RoboMal Dataset}\label{sec:dataset}
The RoboMal dataset (\url{https://doi.org/10.4231/YN7G-H807}) was created to facilitate research in the area of detecting malware from binary executable for not only robotic systems, but also simpler embedded actuator-based CPS. We consider the paradigm of small-scale autonomous cars network as shown. These cars have a variety of sensors, such as LiDARs, ultrasonic sensors, and cameras. They also have motors as actuators. The F1/10 car is one such example, as shown in Fig. \ref{fig:robomal_car_datset}(a). The controller of such cars manages both steering and velocity control. These cars are being increasingly used for education, research, and even racing purposes. Hence, we developed the dataset using the controller files of one such publicly available autonomous car \cite{o2019f110}. The provided F1/10 code repository was updated and modified for this analysis.

The RoboMal dataset uses the wall following controller for the small-scale autonomous cars. The track of the car is shown in Fig. \ref{fig:robomal_car_datset}(b). To inject malicious behavior, variables such as gains and scalars were modified in a way that the performance of the car degraded drastically. We aimed to see clear deviation from expected behavior of the car. We also manipulated the Proportional–Derivative (PD) control structure to generate malware files. All these approaches are potential methods that hackers can deploy to inject malware in such devices. The Gazebo-based simulation was used to ascertain the behavior of the cars with the modified code. Also, good code samples were created by tweaking the parameters in a way that the change in performance was within acceptable bounds. Notably, one realizes that there is no explicit definition of malware in such systems as the decision boundary can be fuzzy. The intention of the program matters as poor performance could just be a result of sub-optimal programming. Keeping this in mind, we utilized multiple samples created by different programmers to build this dataset. 

The distribution of the dataset is shown in Fig. \ref{fig:robomal_car_datset}(c). A total of 450 binary executable ELF files are provided in this dataset with 232 malware files and 218 good software files. The malware files are the ones that showed erratic behavior such as the car crashing into the wall, car not moving, car moving in the wrong direction, etc. On the other hand, good software files result in expected performance but the different values of gains can interfere with the speed and the overall time the car takes around the track. Each file has different changes and these changes have been documented in the detailed label excel file included with the dataset. Binary executables are language-agnostic and are widely used in detecting malware in the case of Windows and Android platforms. Binary executables occupy less memory and are independent software with all necessary libraries already packed. Hence, they are atomic units that can be analyzed without additional resources. The value of this dataset lies in the fact that it provides a distribution of samples for both good and malicious software for an embedded system. 

\section{The RoboMal Framework}\label{sec:model}

Static analysis of code requires identification of patterns in the binary executable, rather than relying on the behavior of the code at runtime. In robotic systems, the ability to stop the execution of malware is particularly useful since a malware might cause life-threatening harm.

\begin{figure*}[hbt!]
  \centering
  \includegraphics[width=0.70\textwidth]{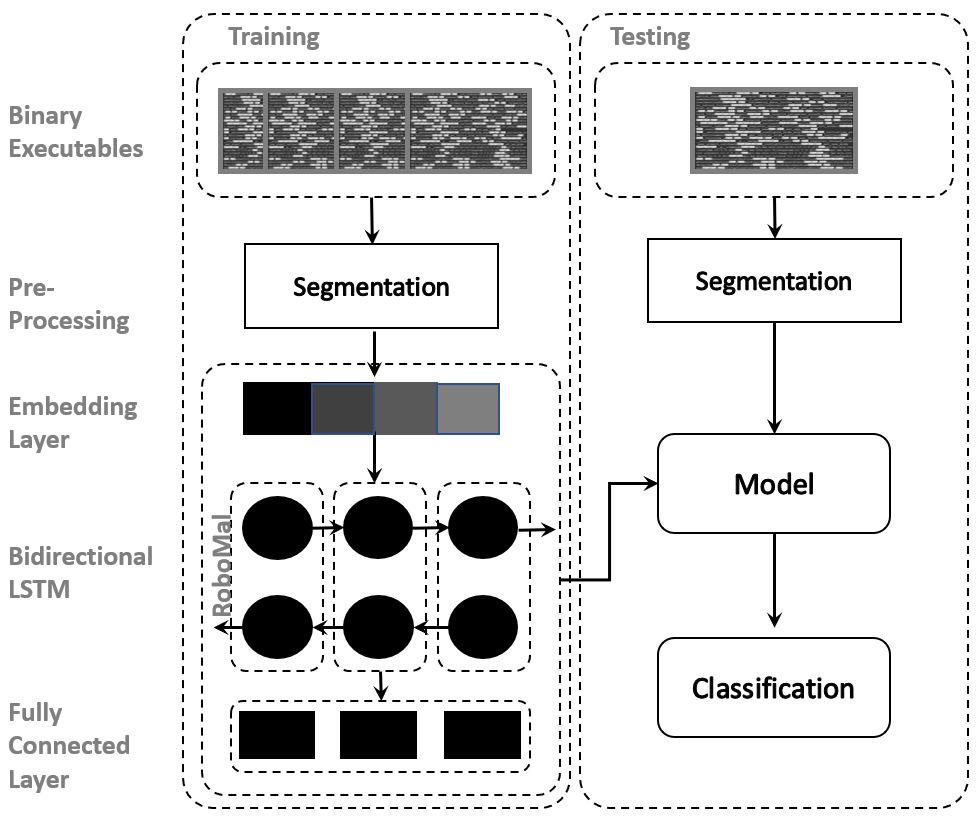}
  \caption{The flow diagram of the RoboMal framework.}
  \label{fig:RoboMal_framework_diagram}
\end{figure*}

In this model, we use the dataset $D$ with $n$ number of samples ${[D_{1},D_{2},...,D_{n}]}$, each belong to class malware (label = 1) or good software (label = 0). Each sample $D_{i}$, where $i\in (1,n)$, represents a raw binary executable, which is then parsed for identification of malware. The workflow of the entire RoboMal framework is shown in Figure \ref{fig:RoboMal_framework_diagram}.  
\subsection{Pre-Processing}
Since the computing power on robotic systems is very limited, we would like to minimize the size of our malware detector. The ELF files of the robot controller are around 18 MB, which hinders the practicability to directly apply deep learning to the raw bytes, especially sequence-based models, for malware detection. Therefore, we pre-processed the executables in order to reduce the input size based on the following two intuitions. First, the controlling logic of the robot only occupies a small proportion of the executable file, while the rest are mostly library and utility functions. Second, since the controller programs are for the same model of robot, we expect that the majority of the variation in the ELF files reside in only a small proportion of the file, which corresponds to the script that describes the controller logic. Therefore, if we know the address of the controller script in the ELF, we can extract the bytes that are projected from the controller script which greatly reduces the input size to the learning models. 

To implement our solution, we first used the $readelf$ command line tool to identify the address of each section header, out of which, we are interested in the $.pydata$ section that covers the controller script written in Python. Then, we extract the hexadecimal version of the file. We observed that the parts corresponding to the controller scripts always have the same starting address, but vary in lengths within the range of 1,000 to 1,300 bytes. Finally, based on the starting address and the compressed file size that we identified, we dumped the binary at the corresponding proportion of the ELF files using $xxd$ command. To make sure our analyses covers the entire program, we capped the length that we extract to 2,000 bytes. We regard each unique byte value as one categorical value. Thus, the pre-processed malware can be viewed as a 2000-word sequence that has a vocabulary size of 256.

\subsection{Long Short-Term Memory}
LSTM models are a type of recurrent neural networks which possess the ability to retain short term memory. They are capable of learning complex and inter-related input features and predict output sequences \cite{malhotra2016lstm}. LSTMs are useful in processing long sequences as they are able to comprehend and retain the overall global distribution as well as map that to local changes. 

In the malware detection application, we input the dataset feature stack $X$ during the training phase. The model first adds an embedding around the input and processes the sequence using the bidirectional LSTM layer. LSTMs usually operate on the time axis, but here we adapt them on the feature space. The bidirectional LSTM captures both past context as well as future context, thereby enabling it to fully comprehend the similarities and differences of the sequences of each binary executable. Given a sequence $X = \{x_{1},x_{2},x_{3},...,x_{l}\}$ for a set of length $l$, the model with memory $c_{t}$ and hidden state $h_{t}$ are defined as:

\begin{equation}
    \centering
 \left[ \begin{smallmatrix} i_{t}\\ f_{t}\\ o_{t}\\ c_{t} \end{smallmatrix}   \right] = \left[ \begin{smallmatrix} \sigma\\ \sigma\\ \sigma\\ tanh \end{smallmatrix}   \right] W [h_{t-1},x_{t}],
  \label{eq:1}
\end{equation}
 where $i$, $f$, and $o$ are the input gate activation, forget gate activation, and output gate activation, respectively \cite{zhou2016text}. 
 The LSTM layer is connected to a fully connected neural network layer with sigmoid activation,
 
\begin{equation} 
y\hat {} = sigmoid(W_{f}M^\prime +b_{f}),
\end{equation}
where $W_{f}$ and $b_{f}$ are the weights of the fully connected layer. The loss function used for this model is binary cross entropy, defined as
\begin{equation} 
loss = - \sum _{i=1}^{e} y_{i}log(y\hat {}_{i} \thinspace).
\end{equation}

Notably, in the code the sigmoid activation layer and binary cross entropy loss is combined into a single BCEwithLogitLoss layer.

\subsection{Comparison with Baselines}
The task of learning the distribution of malware and good software in the RoboMal dataset is primarily an exercise in understanding the distribution of key sequences in the executable. Hence, to evaluate the performance of the RoboMal framework, we present three baselines using Gated Recurrent Units (GRUs), Convolutional Neural Networks (CNNs), and Artificial Neural Networks (ANN) models. GRUs are another form of recurrent neural networks (RNNs) and unlike LSTMs, their retention and removal of memory is controlled by a single gate. GRUs also lack a cell state. While GRUs are faster, they tend to overfit the dataset more than LSTMs. 

CNN are widely used for many machine learning applications, and are good at finding patterns in a file or sequence. While they are good at understanding the local distribution, they falter in mapping the global distribution of the dataset. Further, fully connected deep neural networks can be used for processing sequences, but they have not achieved state-of-the-art performance in most cases. 

\subsection{Evaluation Metrics}
The performance of the models was measured based on the metrics: accuracy, precision, recall, f1 score, false positive rate, and false negative rate. Accuracy measures the ratio of correct predictions over the total number of samples evaluated. It is denoted as:
\begin{equation} 
\centering
{Accuracy}=\frac {tp + tn} {tp+fp+tn+fn},
\end{equation}
where $tp$, $tn$, $fp$, and $fn$ are the number of true positives, true negatives, false positive and false negatives samples, respectively. Precision is the measure of correctly classified true positive samples from all the samples classified as positive,
\begin{equation} 
\centering
{Precision}=\frac {tp} {tp+fp}.
\end{equation}

Recall is the measure of positive samples in all correctly classified samples, 
\begin{equation}
    {Recall}=\frac {tp} {tp+fn}.
\end{equation}

F1-Score is harmonic means between recall and precision. The False Positive Rate (FPR) is ratio between the incorrectly classified negative samples to the total number of negative samples and False Negative Rate (FNR) is the measure of positives samples that were incorrectly classified:
\begin{equation}
    {F1~Score}=\frac {2 \times Precision\times Recall} {Precision+Recall},
\end{equation}
\begin{equation}
    {FPR} = \frac {fp} {fp + tp},
\end{equation}
\begin{equation}
    {FNR} = \frac {fn} {fn + tp}.
\end{equation}

\section{Experiments and Results}\label{sec:results}

\subsection{Experimental Setup}
The RoboMal dataset was used for the validation of the performance of the RoboMal framework. Using a 10-fold cross-validation, the dataset was split into training and test sets. The training and testing were conducted on a PC with NVIDIA 1080Ti GPU, Intel i7 processor, 16GB RAM, and 500TB memory.

The RoboMal framework uses a bidirectional LSTM with an embedding layer of size 16 units. The hidden layer has 16 units (8 each way), and the number of classes is two. The vocabulary size is 256 characters. The Adam optimizer was used with learning rate of 0.001. Binary cross-entropy with logit loss was used as the loss function. The training was completed in batch size of 36 with 100,000 total maximum steps (therefore, epochs = 100000/36) for each fold.

The GRU model has an embedding layer of 16 units with a hidden layer with 16 units and dropout of 0.30. The Adam optimizer with learning rate of 0.001 was used. The binary cross-entropy with logit loss was used as the loss function. The Adam optimizer was used with learning rate of 0.001. The training was completed in batch size of 44 with 100,000 total maximum steps (therefore, epochs = 100000/44) for each fold. The CNN model uses three 1D convolutional layers with 8/16/16 units and stride set as 1. A batch normalization layer was used after each convolutional layer along with dropout of 0.20. The fully connected layers with 480 input size was used as an output layer with sigmoid activation. In this case, we use the binary cross entropy loss and the Adam optimizer (learning rate = 0.001, weight decay = 0.003). The batch size was set to 32 with a maximum step size of 50,000 (epochs = 50000/32) for each fold. 

The ANN model uses three fully connected layers with input dimensions of 2000/200/200. The first two layers use rectified linear units (ReLU) activation with the final activation being sigmoid. We use the binary cross entropy loss and the Adam optimizer (learning rate = 0.001, weight decay = 0.003). The batch size was set to 32 with a maximum step size of 50,000 (epochs = 50000/32) for each fold.

\subsection{Results}
\begin{figure}
  \centering
  \includegraphics[width=1.0\linewidth]{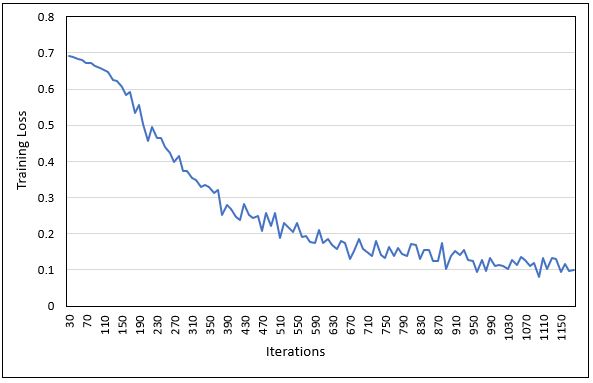}
  \caption{The training loss for RoboMal framework with number of epochs.}
  \label{fig:training_loss_plot}
\end{figure}

The models were compared based on their overall performance on the RoboMal dataset. Table \ref{tab:results} shows the average performance metrics of the models in 10-fold cross validation on the RoboMal dataset. The RoboMal framework outperforms all other baseline models. This performance can be attributed to the bidirectional LSTM as it holds both past and future predictions while making the decision. The training loss for the model is shown in Fig \ref{fig:training_loss_plot}. Also, while GRU model did better than the CNN and ANN models, the lack of independence of gates did not help it achieve great performance. Hence, we can conclude that while processing sequences, the RoboMal framework was able to comprehend the similarities in the samples as well as identify the specific sequences that changed. 

 \begin{table}[]
\centering
\caption{The performance results of the models on the RoboMal Dataset.}
\label{tab:results}
\scalebox{1.0}{
\begin{tabular}{|l|l|l|l|l|l|l|}
\hline
\multirow{2}{*}{Model} & \multicolumn{6}{c|}{Metrics}                           \\ \cline{2-7} 
                       & Accuracy & Precision & Recall & F1 & FPR  & FNR  \\ \hline 
GRU                    & 0.78     & 0.81      & 0.83   & 0.83     & 0.17 & 0.19 \\ \hline
CNN                    & 0.76     & 0.79      & 0.73   & 0.75     & 0.19 & 0.27 \\ \hline

ANN                    & 0.70     & 0.71      & 0.70    & 0.71     & 0.31  & 0.30  \\ \hline

RoboMal                    & \textbf{0.85}     & \textbf{0.87}      & \textbf{0.87}   & \textbf{0.87}     & \textbf{0.13} & \textbf{0.17} \\ \hline
\end{tabular}}
\end{table}

Further, CNN model achieved an accuracy of 0.76 with a precision of 0.79. The model showed balanced behavior with a false positive rate of 0.19 and false negative rate of 0.27. The convolutional filters in this case caused loss of granularity of information which might have resulted in the observed performance. Also, CNN fails to learn global distribution which could also be responsible for the sub-optimal performance. The ANN model achieved an accuracy of 0.7 which while lower than other models, highlights that the distribution of the dataset is enough for all the models to generalize and find decision boundaries. In this case, the false positive and false negative rates were higher, 0.31 and 0.30, respectively. This could be due to the simplicity of the model. 

Hence, we conclude that for static analysis of binaries, the model needs to understand the local distribution of characters and map to the global distribution in the dataset. Also, the baselines show the generalizability on the dataset. 

\section{Conclusion}\label{sec:con}
In this paper, we address the need for malware detection in robotic systems by providing the novel RoboMal dataset and the RoboMal framework. The RoboMal dataset is a collection of 450 binary executables (232 malware files and 218 good software files) of controllers of small-scale autonomous cars. This publicly available dataset is truly unique since it is the first of its kind, with validated working controller code and malicious code executables.

The RoboMal framework analyzes patterns in static binary executables using a bidirectional LSTM-based model with embedding for identifying the maliciousness of the code. The framework's performance was compared against that of other state-of-the-art models, such as GRU, CNN, and ANN. The RoboMal framework outperforms all models with an accuracy of 85\% and precision of 87\%. Notably, this performance was achieved with a dataset of just 450 files and still resulted in 13\% false positive rate and 17\% false negative rates. This shows that the model can generalize on the dataset and is not overfitting the samples. Further, both GRU and CNN models also resulted in an acceptable performance on the RoboMal dataset, thereby proving the generalizability of the distribution of samples in the dataset. Hence, we prove the effectiveness of the framework on the RoboMal dataset.

In the future, we plan to further extend the RoboMal dataset to enable more robust performance. The RoboMal framework can be extended to more semi-supervised and unsupervised approaches using the presented dataset. Further, malware isolation strategies can be created and evaluated for robotic networks by augmenting this framework. 


\section*{Acknowledgment}
The authors acknowledge the support of USDA Grant 2018-67007-28439 in the fulfillment of this work. The authors also acknowledge the help of Mythra VBS Balakunthala in providing the initial setup of the autonomous rally cars. We acknowledge the support of Shih Huan Hung in the development of the RoboMal dataset.



%
\bibliographystyle{IEEEtran}
\bibliography{main}

\end{document}